\documentclass[10pt,twocolumn,letterpaper]{article}

\usepackage{cvpr}              
\definecolor{cvprblue}{rgb}{0.21,0.49,0.74}
\usepackage[pagebackref,breaklinks,colorlinks,allcolors=cvprblue]{hyperref}

\def\paperID{} 
\def\confName{CVPR}
\def\confYear{2026}

\title{Grab-3D: Detecting AI-Generated Videos from 3D Geometric Temporal Consistency}

\author{
Wenhan Chen \qquad Sezer Karaoglu \qquad Theo Gevers\\
University of Amsterdam\\
{\tt\small \{wh.chen, s.karaoglu, th.gevers\}@uva.nl}
}

\usepackage{booktabs}
\usepackage{multirow}
\usepackage{graphicx}
\usepackage{caption}
\usepackage{subcaption}

\begin{document}
\maketitle
\hyphenation{geometric}
\begin{abstract}
\it
Recent advances in diffusion-based generation techniques enable AI models to produce highly realistic videos, heightening the need for reliable detection mechanisms. However, existing detection methods provide only limited exploration of the 3D geometric patterns present in generated videos.
In this paper, we use vanishing-points as an explicit representation of 3D geometry patterns, revealing fundamental discrepancies in geometric consistency between real and AI-generated videos. We introduce \mbox{Grab-3D}, a geometry-aware transformer framework for detecting AI-generated videos based on 3D geometric temporal consistency.
To enable reliable evaluation, we construct an AI-generated video dataset of static scenes, allowing stable 3D geometric feature extraction. We propose a geometry-aware transformer equipped with geometric positional encoding, temporal–geometric attention, and an EMA-based geometric classifier head to explicitly inject 3D geometric awareness into temporal modeling. 
Experiments demonstrate that Grab-3D significantly outperforms state-of-the-art detectors by 11.8\% AP and 14.9\% F1-score, achieving robust cross-domain generalization to unseen generators.

\end{abstract}    
\section{Introduction}
\label{sec:intro}

\begin{figure}[t]
  \centering
  \includegraphics[width=1.0\linewidth]{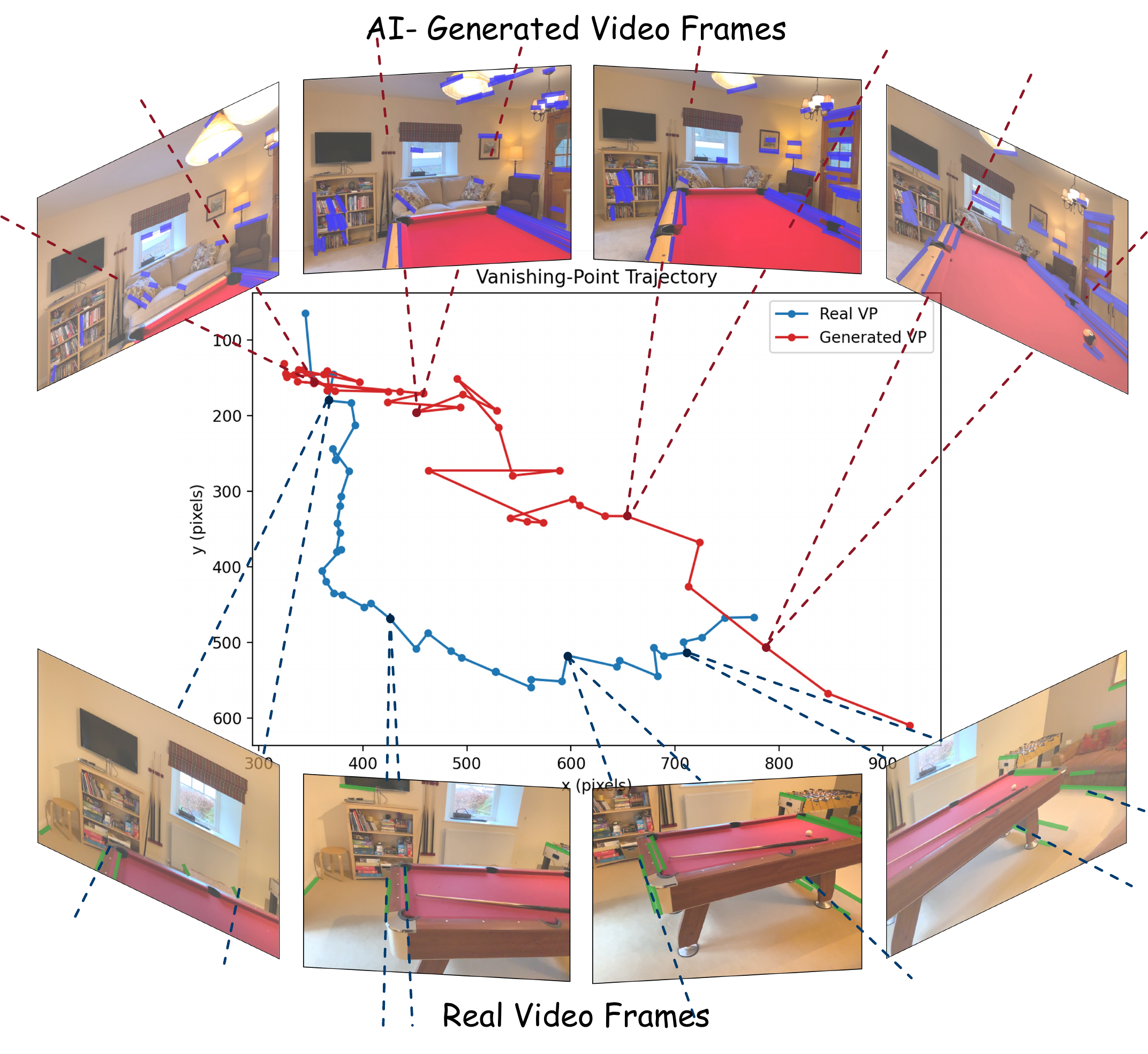}
  \caption{Comparison of vanishing-point trajectories between AI-generated and Real videos.  Real video (\textbf{blue}) shows smooth and consistent motion, whereas generated video (\textbf{red}) shows an unstable and jittery trajectory.}
  \label{fig:introduction}
\end{figure}

With the advancement of AI-generated content technologies, generative models have become widely adopted tools in digital content production. However, as video generation models~\cite{sora2025openai, wan2025wan} gain widespread application, their realism and temporal coherence have made it increasingly difficult to distinguish them from real videos. This poses significant challenges to the authenticity and security of media information~\cite{barrett2023identifying}. Therefore, the demand for efficient and reliable video detection methods has become increasingly urgent.

Existing AI-generated video detectors primarily leverage texture structure~\cite{khan2021video}, spatial artifacts~\cite{zheng2025d3}, or RGB motion patterns~\cite{lin2024detecting}, but these features tend to be brittle across generators and content variations. Some transformer-based approaches introduce temporal modeling to capture relations between 2D features, yet they still overlook how 3D geometry should evolve in real scenes.

These limitations motivate this paper: \textit{Can we introduce 3D consistency into AI-generated video detection tasks?} In real videos, the 3D features and geometric structure of the scene remain physically consistent over time. In contrast, AI-generated videos exhibit deviations from such consistency due to imperfect 3D simulation~\cite{chang2024far, ren2025gen3c}. 
To represent 3D geometry explicitly, we use vanishing-point as a physically grounded geometric descriptor.
Our empirical findings show that the vanishing point in real videos follows a stable motion pattern, whereas AI-generated videos exhibit significant fluctuations (Fig.~\ref{fig:introduction}).  

To ensure stable 3D representations and avoid interference from object motion during feature extraction, we construct a video dataset of static scenes that contain only camera motion, with no moving objects. The dataset contains 5,542 real and AI-generated videos produced by multiple mainstream video generation models~\cite{blattmann2023stable,sora2025openai,veo2025google,runway2025gen4}, covering diverse scenes. Each generated video follows the geometric and semantic conditions of its corresponding real video, with prompt constraints to maintain scene staticity.

We propose Grab-3D, a geometry-aware transformer framework for detecting AI-generated videos by modeling geometric temporal consistency. To enable \mbox{Grab-3D} to learn 3D geometry, we introduce three functional components: Geometric Positional Encoding, which encodes temporal cues through geometric differences; Temporal-Geometric Attention, which incorporates inter-frame geometric similarity into attention; and Geometric Classifier Head, which captures temporal geometric instability via EMA-based residuals. We further evaluate Grab-3D's cross-domain detection performance and demonstrate its superior generalization compared with existing methods. The results show that Grab-3D achieves state-of-the-art performance, demonstrating the effectiveness of modeling 3D geometric temporal consistency for AI-generated video detection.

Our contributions can be summarized as follows:

\begin{itemize}[leftmargin=*, label=\textbullet]
    \item We present an innovative study of 3D geometric temporal consistency in AI-generated videos, demonstrating 3D geometric differences between real and synthetic videos using vanishing–point–based cues.
    \item We construct a static AI-generated video dataset, enabling stable extraction of 3D geometric features and practical evaluation of detection performance.
    \item We propose Grab-3D, a geometry-aware transformer framework equipped with geometric positional encoding, temporal–geometric attention, and geometric classifier head, explicitly injecting 3D geometric awareness into temporal modeling. 
\end{itemize}
\section{Related Work}
\label{sec:relatedwork}

\subsection{Video Generation Models}

With the advent of diffusion models and their success in image synthesis, diffusion-based video generation rapidly became the dominant approach~\cite{zhang2023i2vgen,chen2023videocrafter1,chen2023seine}. Large-scale generative models further advanced transformer-based video diffusion architectures~\cite{ho2022video,lu2023vdt,ma2024latte,chen2024gentron}, achieving unified spatio-temporal modeling rather than frame-wise generation or independent denoising. Compared with conventional video generation models, these architectures exhibit stronger temporal reasoning and logic coherence. The combination of large language models (LLMs)~\cite{kondratyuk2023videopoet,chen2023videollm,maaz2023video} enables text–video alignment, enhancing semantic understanding for both text-to-video and image-to-video generation. 
Although video generation models can produce realistic and coherent videos, they still fail to maintain geometric patterns of real world. Our work aims to investigate 3D geometric discrepancies between generated and real videos.


\subsection{AI-generated Video Detection}

Due to the rapid development of DeepFake technologies and their social implications, early research on synthetic video detection focused on face-manipulated videos~\cite{pei2024deepfake}. These methods exploit local texture artifacts~\cite{ba2024exposing,cao2022end,shiohara2022detectingdeepfakesselfblendedimages} and noise distribution~\cite{wang2023noise} to identify spatial inconsistencies, while physiological features~\cite{peng2024deepfakes} and in-frame differences~\cite{choi2024exploiting,xu2024learning} are used to detect time anomalies. But when applied to general AI-generated videos beyond Deepfake, these approaches become severely limited~\cite{cao2023comprehensive}.

With the evolution of video generation techniques, the visual fidelity and spatio-temporal consistency of synthetic videos have become increasingly realistic, making low-level artifact cues less reliable. This shifts the focus of the detection toward higher-level structural and consistency cues. Ma et al.~\cite{ma2024detecting} propose DeCoF, which models frame-wise consistency to capture temporal artifacts. Chen et al.~\cite{chen2024demamba} construct a million-scale GenVideo dataset and introduced DeMamba, a structured state-space model based on Mamba architecture~\cite{gu2024mamba}. UNITE~\cite{kundu2025towards} leverages SigLIP to extract domain-agnostic features and introduces an attention-diversity loss. Yang et al.~\cite{yang2025d} adopt discrepancy learning under multi-generator training to extract generator-agnostic artifacts. Chang et al.~\cite{chang2024far} combine semantic, optical flow, and depth features, revealing geometric and motion inconsistencies between real and generated videos.

However, despite these advances, existing detectors still operate primarily on 2D appearance features and lack explicit modeling of 3D geometric consistency, a fundamental physical prior in real videos. This gap motivates our geometry-aware temporal framework.


\subsection{3D Geometric Consistency}

In authentic videos, scenes follow stable 3D geometric constraints, where the depth, motion, and perspective relationships across frames obey the physical laws~\cite{naseer2018indoor}. In contrast, current video generation models lack reliable 3D understanding. Both text-to-video and image-to-video models generate sequences conditioned only on text or a single frame, making it difficult for them to maintain consistent 3D spatio-temporal relationships~\cite{chang2024far}.

Several works aim to enhance 3D awareness of generative models by introducing geometric consistency. 
Höllein et al.~\cite{hollein2024viewdiff} propose ViewDiff, which combines cross-view attention and 3D projection layers within the diffusion, enabling 3D object rendering from arbitrary viewpoints.
Xu et al.~\cite{xu2024camco} enhance 3D-consistent video generation with camera ego-motion and dynamic objects by introducing epipolar-constraint attention and camera parameterization.
GEN3C~\cite{ren2025gen3c} leverages an explicit 3D cache by unprojecting depth maps into 3D space, improving temporal consistency and geometric coherence with 3D representation. 
Zhang et al.~\cite{zhang2025world} propose World-consistent Video Diffusion, which provides explicit supervision for video generation by learning 3D geometric structure and appearance rendering.
These approaches collectively demonstrate a trend toward integrating explicit or implicit 3D features into models to improve geometric consistency across views and time. 

\section{3D Feature Extraction}
\label{sec:3dfeatureextraction}

Inspired by recent 3D-aware generative models, we introduce the concept of 3D geometric consistency into the video detection task. 
Our method focuses on \textbf{vanishing-point-based representations} as explicit and physically grounded descriptors of 3D scene geometry~\cite{481545,lim2022uv}. 
In this section, we first describe the vanishing-point extraction method, followed by the geometric representation used in our framework.



\paragraph{Intrinsic Parameter Estimation.}
To ensure the accuracy of each 3D geometric feature, we assign every frame a calibrated intrinsic matrix $K_t$. 
For real videos, ${K}_t$ is calculated from EXIF metadata or calibration sequences. 
For generated videos, we leverage their one-to-one correspondence with real videos and adopt a paired intrinsic compensation strategy~\cite{yu2022roma, hagemann2023deep}: a reference intrinsic ${K}_{\text{ref}}$ is computed as the median of all per-frame intrinsics from the corresponding real video and then rescaled to match the resolution of the generated one. For each generated frame, the compensated intrinsic is defined as:
\begin{equation}
{K}_{\text{ref}}^{\text{gen}} = s_h s_w {K}_{\text{ref}},
\end{equation}
where $s_w$ and $s_h$ are the scaling factors between the real and generated resolutions. 
All subsequent geometric computations are performed under these calibrated intrinsics.

\paragraph{Vanishing Point Detection.}
We employ the open-source library \text{luvpdetect}~\cite{lu20172} for single-frame vanishing-point estimation. Candidate line segments are detected using the Hough transform and grouped by orientation (Fig.~\ref{fig:dataset}.c).   
RANSAC-based clustering is then used to estimate the three dominant vanishing
directions $(\mathbf{v}_1^{2D}, \mathbf{v}_2^{2D}, \mathbf{v}_3^{2D})$, 
corresponding to the projections of mutually orthogonal axes in the 3D scene.
Each detected 2D vanishing-point $\mathbf{v}_i^{2D}=(u_i, v_i)$ is unprojected into a normalized 3D direction using the per-frame camera intrinsics $K_t$:
\begin{equation}
\mathbf{v}_i^{3D} = 
\frac{K_t^{-1}[u_i, v_i, 1]^\top}
{\|K_t^{-1}[u_i, v_i, 1]^\top\|}.
\label{eq:vp_projection}
\end{equation}
The resulting $\{\mathbf{v}_1^{3D}, \mathbf{v}_2^{3D}, \mathbf{v}_3^{3D}\}$ 
encode orthogonal 3D directions under the camera coordinate system.

\paragraph{Cross-domain Normalization.}
We express all vanishing directions
in a unified reference camera coordinate system defined by the per-video calibration
matrix $K_{\text{ref}}$ (rescaled to the working resolution).

For real videos, we compensate frame-wise intrinsic fluctuations by mapping each direction into the
reference camera space. The normalized 3D direction is defined as:
\begin{equation}
\mathbf{v}_i^{3D} = 
{K}_{\text{ref}}^{-1} {K}_t \mathbf{v}_i^{2D}.
\label{eq:real_norm_ref}
\end{equation}

For generated videos, which lack valid per-frame intrinsics, we directly unproject
each vanishing point using the paired reference intrinsics:
\begin{equation}
\mathbf{v}_i^{3D} =
{K}_{\text{ref}}^{-1} \mathbf{v}_i^{2D}.
\label{eq:gen_norm_ref}
\end{equation}
This assumes that the generated video is rendered under the same virtual
camera as its real counterpart, enabling cross-domain geometric alignment.


\paragraph{Temporal Geometric Representation.}
For each frame $t$, we concatenate the detected vanishing directions and associated geometric cues into a 21-dimensional feature vector:
\begin{equation}
f_t = 
[\mathbf{v}_1^{3D}, \mathbf{v}_2^{3D}, \mathbf{v}_3^{3D},
 \mathbf{v}_1^{2D}, \mathbf{v}_2^{2D}, \mathbf{v}_3^{2D},
 d_t, m_t] \in \mathbb{R}^{21},
\label{eq:vp_feature}
\end{equation}
where $\mathbf{v}_i^{3D}$ and $\mathbf{v}_i^{2D}$ represent the 3D and 2D vanishing
directions, $d_t$ encodes the outside distance of each vanishing point, and
$m_t$ is the visibility mask; both $d_t$ and $m_t$ are used for quality filtering.
We also record the set of supporting line segments $L_t$ associated with each vanishing-point obtained from the RANSAC-based clustering process.

The per-frame features are then stacked along the temporal dimension to form
the sequence:
\begin{equation}
F = [f_1, f_2, \dots, f_T] \in \mathbb{R}^{T\times 21},
\label{eq:temporal_input}
\end{equation}
which serves as the geometric input sequence for the Transformer-based backbone described in Sec.~\ref{sec:geometryawaretransformer}.

\section{Video Dataset}
\label{sec:staticvideodataset}

\begin{figure*}[t]
  \centering
  \includegraphics[width=1.0\linewidth]{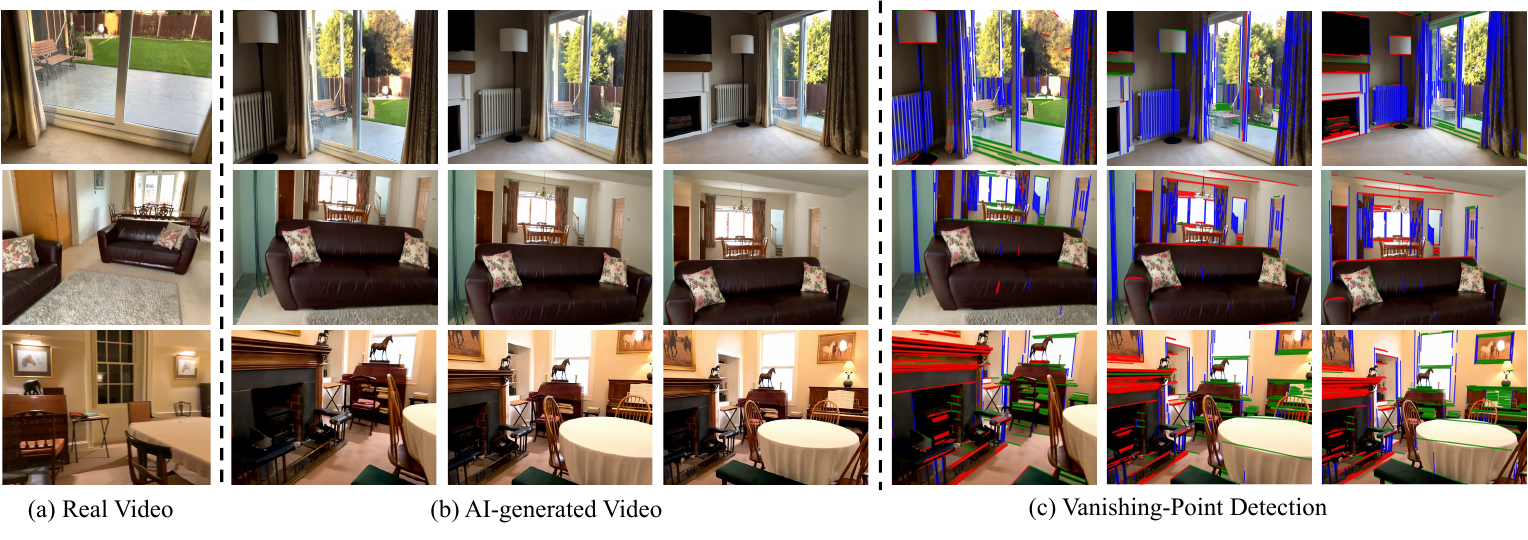}
  \caption{Overview of the video dataset of static scenes. (a)Real videos, (b)Paired AI-generated Videos produced by Wan2.2, (c)Vanishing-point detection results obtained by clustering detected line segments. Red, blue, and green lines represent the supporting lines corresponding to the three vanishing points.}
  \label{fig:dataset}
\end{figure*}

Existing AI-generated video datasets, such as DeManba~\cite{chen2024demamba}, GenVidBench~\cite{ni2025genvidbench}, and IVY-FAKE~\cite{zhang2025ivy}, mainly contain scenes with dynamic foregrounds and relatively static backgrounds. However, the presence of moving objects introduces challenges for extracting reliable 3D geometric features. Vanishing-point computation relies on the stable geometric structure of static scenes, whereas edges on moving objects violate global geometric constraints and move independently over time~\cite{wang2024survey,qin2022rgb,li2024learn}. 

To fairly evaluate the effectiveness of 3D consistency in video detection, we construct a \textbf{video dataset of static scenes} where all scenes are free of moving objects (\eg, humans, animals, or vehicles)with only camera motion ensuring that the underlying scene geometry remains consistent across frames. 

\paragraph{Real-world Videos.}
To ensure that all real videos remain completely static, we collected samples from widely used 3D reconstruction and SLAM datasets, such as Arkitscenes~\cite{baruch2021arkitscenes} and NYUdepth~\cite{Silberman:ECCV12}, which capture static scenes with controlled camera motion (Fig.~\ref{fig:dataset}.a). In total, 2,284 real static video clips are collected. 

\paragraph{AI-generated Videos.}
 To construct a generated video dataset (Fig.~\ref{fig:dataset}.b) aligned with the real static videos, we employ three open-source generative models, \textit{Wan 2.2}~\cite{wan2025wan}, \textit{CogVideoX}~\cite{yang2024cogvideox}, and \textit{Stable Video Diffusion (SVD)}~\cite{blattmann2023stable}, covering both text-to-video (T2V) and image-to-video (I2V) generation models. To ensure geometric comparability under the same scene structures,  generation conditions are strictly derived from the corresponding real videos. We use the large language model \textit{Qwen}~\cite{bai2025qwen2} to generate descriptive prompts for each real video. During this process, we constrain the prompt content to describe only static elements such as architecture, furniture, and objects.  For T2V models, the textual prompt is directly used as input, while for I2V generation, the first frame of the corresponding real video is also used as input. Each prompt is further appended with explicit constraints, such as \textit{“Static scene with no people, no object movement, only camera moving”} to suppress any dynamic content and maintain a static scene. 

In addition, for several mainstream but not fully open-source commercial models, including \textit{Sora}~\cite{sora2025openai}, \textit{Veo}~\cite{veo2025google}, and \textit{Gen-4}~\cite{runway2025gen4}, we generated a small number of samples using the same method as above. These videos are incorporated in the test dataset to evaluate our model's cross-domain detection under unseen generative conditions.

\paragraph{Dataset Summary.}
As shown in the Table~\ref{tab:dataset}, we constructed the first \textbf{static AI-generated video dataset} through the above pipeline, 
comprising 3,322 paired real–generated videos in total. 
Each synthetic video is generated under the geometric and semantic conditions of its real counterpart, ensuring one-to-one correspondence and scene consistency across domains. 
The dataset covers a range of mainstream generation models, and provides a clean benchmark for evaluating geometric consistency based video detection.

\begin{table*}[t]
\centering
\setlength{\tabcolsep}{6pt} 
\renewcommand{\arraystretch}{0.7}
\small
\caption{Overall statistics of the dataset of Real and AI-generated static videos.}
\label{tab:dataset}
\begin{tabular}{lccccccc}
\toprule
\textbf{Video Sources} & \textbf{Type} & \textbf{Task} & \textbf{Resolution} & \textbf{FPS} & \textbf{Length} & \textbf{Training Set} & \textbf{Testing Set}\\ 
\midrule
ARKitScenes~\cite{baruch2021arkitscenes} & Real & - & 640$\times$480 & 30 & 10s & 1500 & 500  \\
NYU Depth V2~\cite{Silberman:ECCV12} & Real & - & 640$\times$480 & 16 & 5--10s & 255 & 29  \\
\midrule
Wan2.2~\cite{wan2025wan} & Fake & T2V \& I2V & 1280$\times$720 & 16 & 5s & 1513 & 498 \\
CogVideoX~\cite{yang2024cogvideox} & Fake & T2V \& I2V & 720$\times$480 & 8 & 6s & 463 & 46  \\
\midrule
Stable Video Diffusion~\cite{blattmann2023stable} & Fake & I2V & 1024$\times$576 & 8 & 3s & - & 498  \\
Sora2~\cite{sora2025openai} & Fake & T2V \& I2V & 1280$\times$768 & 30 & 4s & - & 121  \\
Runway Gen4~\cite{runway2025gen4} & Fake & T2V \& I2V & 1280$\times$768 & 24 & 4-5s & - & 100  \\
Veo3~\cite{veo2025google} & Fake & T2V \& I2V & 1280$\times$768 & 24 & 6-8s & - & 83  \\
\midrule
\textbf{Total Count} & - & - & - & - & - & \textbf{3667} & \textbf{1875}  \\
\bottomrule
\end{tabular}
\end{table*}

\section{Methodology}
\label{sec:geometryawaretransformer}

With the dataset ready, we propose a transformer-based method that focuses on detection of geometric consistency in AI-generated videos: \textit{Grab-3D}.

\subsection{Framework}
As shown in Fig.~\ref{fig:pipeline}, \textit{Grab-3D} is built upon a geometry-aware transformer backbone, referred to as the \textbf{Geometric-Aware Transformer (GAT)}, which combines temporal representation and geometric awareness.  Instead of treating frames as independent 2D observations, GAT represents each frame using its vanishing-point features and learns to reason about their geometric evolution over time. This design allows the model to perceive how 3D spatial structure changes consistently across frames in real videos, while detecting unstable or non-physical geometric behaviors present in generated ones.

GAT introduces two functional components. 
\textbf{Geometric Temporal Transformer} captures temporal dependencies and encodes geometric relationships through dual attention and positional encoding. 
\textbf{Geometric Classifier Head} refines the representation using dynamic geometric residuals derived from vanishing point predictions.

\subsection{Geometric Temporal Transformer}

We first introduce the \textbf{Geometric Positional Encoding (GPE)} module, which injects geometric evolution cues into the frame representations. 
Then, we describe the \textbf{Temporal Geometric Attention (TGA)}, which jointly captures temporal dependencies and geometric consistency.

\paragraph{Geometric Positional Encoding (GPE)}

\begin{figure*}[t]
  \centering
  \includegraphics[width=1.0\linewidth]{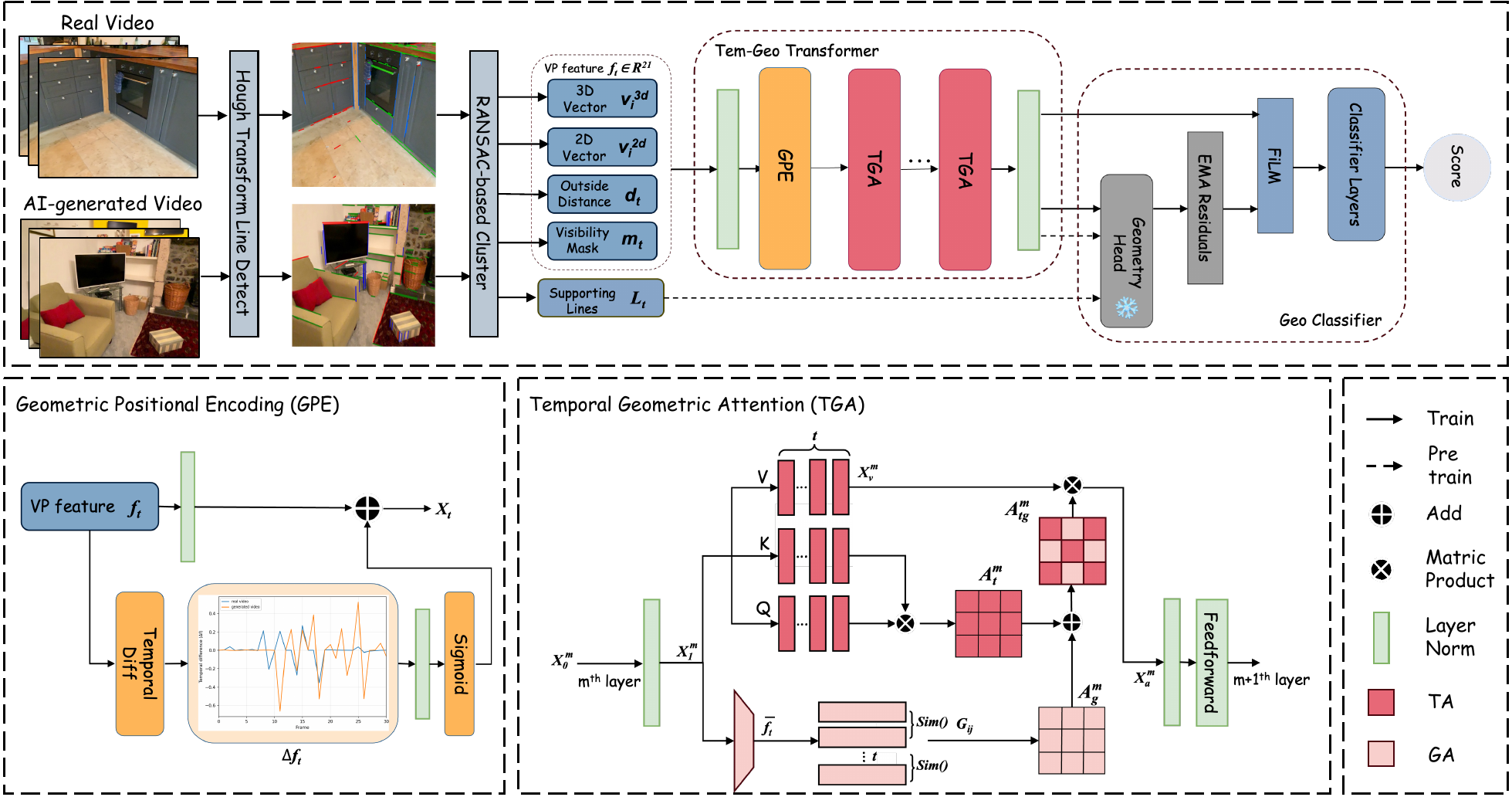}
  \caption{Overall architecture of the Grab-3D, a geometry-aware transformer framework. The model first extracts vanishing-point features from videos, encodes them via the GPE module, and models temporal–geometric dependencies using TGA. A frozen geometry head provides vanishing-point cues for EMA-based geometric residuals before classifier layers.}
  \label{fig:pipeline}
\end{figure*}

Conventional transformers employ sinusoidal or learnable positional embedding to represent the temporal order of frames. 
To make the model sensitive to 3D structural changes, we replace conventional temporal encoding with the \textit{Geometric Positional Encoding (GPE)} based on vanishing-point features, as shown in Fig.~\ref{fig:pipeline}.

Given per-frame geometric features \( f_t \in \mathbb{R}^{21} \), we compute their temporal differences to capture geometric variations between frames, and project them through a linear layer followed by normalization to obtain the geometric embedding:

\begin{equation}
\Delta f_t = f_t - f_{t-1},
\end{equation}
\begin{equation}
E_t = \text{LN}(W_g \Delta f_t),
\end{equation}

where \( W_g \) is a learnable projection matrix.
The initial feature  \( f_t \) is projected to the model dimension using  \( W_f \) and combined with the geometric embedding through a learnable non-negative gate \( \alpha \):
\begin{equation}
X_t =  W_f f_t + \alpha E_t.
\end{equation}

Here, \( \alpha \) controls the strength of geometric modulation and is optimized jointly with the backbone.
This enables the model to perceive temporal order implicitly from 3D geometric changes rather than explicit frame indices. 

\paragraph{Temporal Geometric Attention (TGA)}

To enable transformer layers to reason about temporal and geometric dependencies jointly, we design a \textit{Temporal Geometric Attention (TGA)} that extends conventional self-attention with an additional geometric pathway. 
As shown in Fig.\ref{fig:pipeline}, the GAT is an $M$-layer transformer. In the \(m^{th}\) layer of GAT, the input feature \(X_0^{m}\) of the \(m^{th}\) layer goes through the layer normalization LN:
\begin{equation}
X_1^{m} = \text{LN}(X_0^{m}),\quad m = 1, \ldots, M.
\end{equation}
Then \(X_1^{m}\) is fed into the two parallel branches: temporal attention (TA) calculation and geometric attention (GA) calculation. For the TA branch, we maintain the same structure as standard multi-head self-attention. We use TA to calculate the attention \(A_t^{m}\) and the attention ``value" \(X_v^{m}\) in the \(m^{th}\) layer of GAT. Instead of directly multiplying \(A_t^{m}\) with \(X_v^{m}\), we design the geometric attention module to import the factors of geometry into the attention process.

To incorporate 3D consistency, the GA branch models geometric similarity between frames by injecting geometry-based correlations into the attention map:
\begin{equation}
G_{ij} = \mathrm{Sim}(\bar{f}_i, \bar{f}_j),
\end{equation}
where $\bar{f}_i$ and $\bar{f}_j$ denote the geometric embeddings of the $i$-th and $j$-th frames, and $\mathrm{Sim}(\cdot)$ represents cosine similarity.
In practice, the geometric embeddings are obtained through a learnable projection and normalization, 
$\bar{f}_t = \mathrm{Norm}(W_g f_t)$, 
ensuring $\|\bar{f}_t\|_2 = 1$. 

The resulting similarity matrix $G$ is further scaled by a learnable temperature $\tau > 0$ and normalized with softmax:
\begin{equation}
A_{\mathrm{g}}^{m} = \mathrm{Softmax}\!\left(\frac{G}{\tau}\right).
\end{equation}

The two attention matrices \(A_{\text{t}}^{m}\) and \(A_{\text{g}}^{m}\) are then fused to form the unified temporal–geometric attention matrix:
\begin{equation}
A_{\mathrm{tg}}^{m} = A_{\mathrm{t}}^{m} + \lambda\,A_{\mathrm{g}}^{m},
\end{equation}
where \( \lambda \) is a learnable scalar that balances the contributions of geometric and temporal branches.  
The final output of temporal–geometric attention is obtained as:
\begin{equation}
X_a^{m} = A_{\mathrm{tg}}^{m} X_v^{m}.
\end{equation}
where \(X_v^{m}\) is calculated in the temporal attention branch.

Then, as shown in Fig~\ref{fig:pipeline}, the output \(X_a^{m}\) is passed through layer normalization and feed-forward layers to calculate the final product of the \(m^{th}\) layer of GAT.

\begin{table*}[t]
\caption{Comparison with baselines on Static AI-generated videos, both in-domain and cross-domain. We use \textbf{bold} to indicate the best results and \underline{underlined} to denote the second-best result.}
\label{tab:comparison}
\centering
\setlength{\tabcolsep}{9pt} 
\renewcommand{\arraystretch}{0.8}
\small
\begin{tabular}{l|c|cccccc|c}
\toprule
Detection & Detection & \multirow{2}{*}{Metric} & \multirow{2}{*}{In-domain} & \multicolumn{5}{c}{Cross-domain} \\
\cmidrule(lr){5-9}
Method & level & & & SVD & Sora & Veo & Gen-4 & Avg. \\
\midrule
         &       & AUC & 0.8567 & 0.6573 & 0.7197 & 0.6563 & 0.7816 & 0.7037 \\
STIL~\cite{gu2021spatiotemporal}      & Image & F1  & 0.6841 & 0.4077 & 0.3718 & 0.5626 & 0.5960 & 0.4845 \\
         &       & AP  & 0.8510 & 0.6862 & 0.7021 & 0.6798 & 0.7922 & 0.7151 \\
\cmidrule(lr){3-9}
         &       & AUC & 0.8442 & 0.6593 & 0.6709 & 0.7446 & 0.8284 & 0.7258 \\
NPR~\cite{tan2024rethinking}     & Image & F1  & 0.7814 & 0.5753 & 0.5327 & 0.4932 & 0.7692 & 0.5926 \\
         &       & AP  &  0.8471 & 0.6805 & 0.6600 & 0.7473 & 0.8587 & 0.7366 \\
\midrule
         &       & AUC & 0.9108 & \underline{0.8539} & 0.7817 & 0.7907 & 0.8956 & \underline{0.8304} \\
TALL\cite{xu2023tall}     & Video & F1  & 0.8004 & 0.6861 & 0.5851 & 0.7586 & 0.7543 & 0.6960 \\
         &       & AP  & 0.9178 & \underline{0.8659} & 0.7781 & 0.7920 & 0.8898 & 0.8314 \\
\cmidrule(lr){3-9}
         &       & AUC & 0.9638 & 0.7016 & \underline{0.8601} & 0.7207 & \underline{0.9015} & 0.7965 \\
MINTIME\cite{coccomini2024mintime}     & Video & F1  & 0.9061 & 0.5167 & \underline{0.7611} & 0.7266 & 0.7853 & 0.6974 \\
         &       & AP  & 0.9629 & 0.7119 & \underline{0.8477} & 0.7904 & \underline{0.9090} & 0.8147 \\
\cmidrule(lr){3-9}
         &       & AUC & 0.9167 & 0.7978 & 0.6874 & 0.8366 & 0.8268 & 0.7871 \\
XCLIP\cite{ni2022expanding}  & Video & F1  & 0.6915 & 0.4176 & 0.2319 & 0.7056 & 0.4733 & 0.4571 \\
         &       & AP  & 0.9267 & 0.8154 & 0.7058 & 0.8787 & 0.8477 & 0.8119 \\
\cmidrule(lr){3-9}
         &       & AUC & 0.9346 & 0.8448 & 0.6860 & 0.8555 & 0.8488 & 0.8088 \\
Demamba\cite{chen2024demamba}    & Video & F1  & 0.8365 & \underline{0.7337} & 0.5029 & 0.8302 & 0.7337 & 0.7001 \\
         &       & AP  & 0.9440 & 0.8656 & 0.7177 & \underline{0.8863} & 0.8756 & \underline{0.8363} \\
\cmidrule(lr){3-9}
         &       & AUC & 0.7595 & 0.6880 & 0.7657 & \underline{0.8697} & 0.8250 & 0.7815 \\
D3\cite{zheng2025d3}       & Video & F1  & 0.7131 & 0.6716 & 0.7260 & \underline{0.8448} & \underline{0.7973} & \underline{0.7466} \\
         &       & AP  & 0.7665 & 0.7006 & 0.7848 & 0.8484 & 0.8377 & 0.7876 \\
\midrule
                      &      & AUC & 0.9957 & \textbf{0.9635} & \textbf{0.9248} & \textbf{0.9386} & \textbf{0.9791} & \textbf{0.9515} \\
\textbf{\textit{Grab-3D (Ours) }} & Video & F1  & 0.9726 & \textbf{0.8758} & \textbf{0.7882} & \textbf{0.8516} & \textbf{0.8791} & \textbf{0.8487} \\
                      &       & AP  & 0.9941 & \textbf{0.9695} & \textbf{0.9238} & \textbf{0.9430} & \textbf{0.9799} & \textbf{0.9541} \\
\bottomrule
\end{tabular}
\end{table*}

\subsection{Geometric Classifier Head}

To further exploit geometric cues for classification, we introduce the \textit{Geometric Classifier Head} 
that guides the classification process by temporal geometric residuals. 
We first pre-train a geometry head on real videos using vanishing-point based geometric losses: reprojection consistency \(\mathcal{L}_{\text{repj}}\), temporal smoothness \(\mathcal{L}_{\text{temp}}\), and supporting line segments \(\mathcal{L}_{\text{line}}\):

\begin{equation}
\mathcal{L}_{\text{repj}}
=\sum_{t,i}
\left\|
\pi(\mathbf{u}_i^{t}) - \mathbf{v}_{i}^{2D,t}
\right\|_2^2,
\end{equation}

\begin{equation}
\mathcal{L}_{\text{temp}}
=\sum_{t=2}^{T}\sum_{i}
\left\|
\mathbf{u}_i^{t} - \mathbf{u}_i^{t-1}
\right\|_2^2 ,
\end{equation}

\begin{equation}
\mathcal{L}_{\text{line}}
=\sum_{t,k}
\frac{
\big|\ \ell_k^{t\top}\, \pi(\mathbf{u}_{\,i(k,t)}^{t})\big|}
{\left\|(a_k^t, b_k^t)\right\|_2},
\end{equation}
where \(\mathbf{u}_i^{t}\in \mathbb{R}^{3}\) is the predicted 3D direction of the \(i^{th}\) vanishing-point at frame $t$, $\mathbf{v}_{i}^{2D,t}$ is the ground-truth 2D vanishing-point, and \(\ell_k^t = (a_k^t, b_k^t,c_k^t)\) is the homogeneous line from the set of supporting line segments.

Once pre-trained, the geometry head is frozen and used as a decoder during the main classification training. The output of the GAT is decoded into vanishing-point predictions $U_t \in \mathbb{R}^{3\times3}$. We then compute stateless exponential moving averages (EMA) to model the temporal change of geometric features:
\begin{equation}
\hat{U}_t = \alpha U_t + (1 - \alpha)\hat{U}_{t-1},
\end{equation}
where $\alpha \in [0,1]$ is the smoothing factor. 
Based on $\hat{U}_t$, we derive four types of residuals — angular \(r_t^{ang}\), velocity $r_t^{vel}$, acceleration $r_t^{acc}$, and orthogonality $r_t^{ort}$ — 
that together quantify geometric inconsistency across frames. 

The temporal geometric residuals for each frame $R_t = [r_t^{ang}, r_t^{vel}, r_t^{acc}, r_t^{ort}]\in \mathbb{R}^{4}$ are first projected to the model dimension through a learnable linear layer and then fused with the backbone output $F_t$ via Feature-wise Linear Modulation (FiLM)~\cite{perez2018film}: 
\begin{equation}
[\gamma_t, \beta_t] = W_r R_t,
\end{equation}
\begin{equation}
H_t = F_t \odot (1 + \gamma_t) + \beta_t,
\end{equation}
where $W_r$ is a learnable projection matrix. The fused features $H_t$ are then passed through the classifier head to produce frame-level representations, followed by temporal pooling and a final binary classification to predict logits and confidence scores for real and AI-generated videos.


\section{Experiments}
\label{sec:experiments}

\subsection{Experiment Settings}

As shown in Table~\ref{tab:dataset}, the video dataset is divided into an in-domain set and a cross-domain set. For the in-domain set, we use 3,667 videos for training, 438 for validation, and 439 for testing. The cross-domain set contains 802 pairs of generated and real videos, all of which are used for evaluating the model's out-of-distribution generalization. 

For baselines, we compare our approach with the following AI-generated detection methods, including two image-level detectors, NPR~\cite{tan2024rethinking}, STIL~\cite{gu2021spatiotemporal}, and five video-level detectors, TALL~\cite{xu2023tall}, MINITIME~\cite{coccomini2024mintime}, XCLIP~\cite{ni2022expanding}, Demamba~\cite{chen2024demamba}, and D3~\cite{zheng2025d3}. 

All models in our experiments are trained on NVIDIA RTX A6000 GPUs.
For preprocessing, each video is randomly cropped to a 5-second segment. To ensure consistency with camera intrinsics, all frames are rotated into a horizontal orientation.
We train our model and all baseline models for 100 epochs with a batch size of 32. We use cross-entropy loss to optimize the classifier head and the AdamW optimizer with an initial learning rate of $10^{-4}$.

We evaluate our method using these primary metrics: (1) AUC (Area Under the Precision-Recall Curve); (2) F1-score (confidence threshold = 0.5), and (3) AP (Average Precision). 

\subsection{Detection Evaluation}
To evaluate the detection capability and generalizability of our Grab-3D, we compare it with existing image-level and video-level detectors. 
To ensure a fair evaluation, all baseline models are retrained on our dataset using the same data split and training settings. And we evaluate all methods under both in-domain and cross-domain scenarios. The overall results are summarized in Table~\ref{tab:comparison}.

For in-domain detection, all models exhibit strong detection performance. In particular, all video-level detectors achieve over 90\% AUC and AP.
Note that D3 is training-free, and its in-domain test samples are used in cross-domain evaluation.  
Even image-level detectors, STIL and NPR, achieve AUC and AP over 84\%. Despite this, our method still surpasses all baselines and achieves the highest in-domain performance. 

As for cross-domain detection, we evaluate all models on unseen generators (SVD, Sora, Veo, and Gen-4). As shown in Table~\ref{tab:comparison}, existing detectors experience a performance drop when shifting to cross-domain datasets. 
Image-level detectors struggle the most, with AP dropping by over 15–20\% on average, indicating that image-wise 2D appearance cues are not reliable across video generators. 
Among video-based methods, all baselines exhibit limitations. For instance, TALL and MINITIME show relatively competitive results; however, they are limited by their focus on detecting facial forgeries, making them poorly effective on general AI-generated videos. 
XCLIP and Demamba are both CLIP-based detectors, yet they still suffer from generative paradigms like Sora (70\% and 71\% AUC), suggesting that temporal or appearance modeling is insufficient.
D3 shows stable performance due to its training-free design, but its performance remains lower than ours on all unseen domains. 
In contrast, our Grab-3D demonstrates strong cross-domain generalization even without pre-training on large-scale videos. It still outperforms all baselines across AUC, F1, and AP. On average, Grab-3D surpasses the second-best method by +12.1\% AUC, +14.9\% F1, and +11.8\% AP, confirming that 3D geometric temporal modeling provides robust, generator-agnostic signals.

\subsection{Qualitative Analysis}
\begin{figure}[t]
  \centering
  \includegraphics[width=1.0\linewidth]{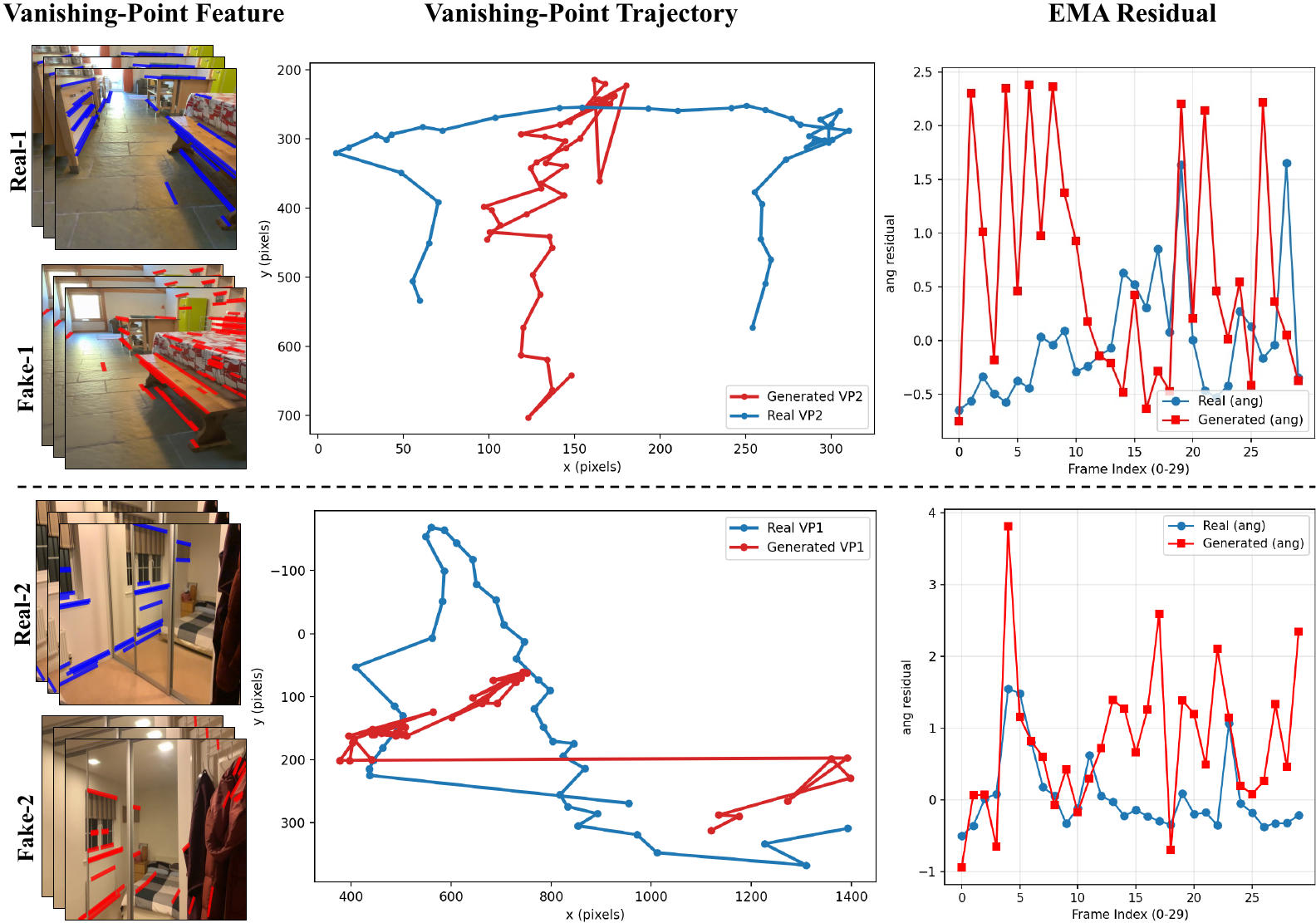}
  \caption{Comparison of vanishing-point trajectories and EMA residuals between real videos (\textbf{blue}) and generated videos (\textbf{red}).}
  \label{fig:analysis}
\end{figure}

We further analyze how Grab-3D captures the 3D geometric differences between real and generated videos by visualizing the projected trajectories of vanishing-point motion on the image plane, as well as the corresponding frame-wise EMA residual sequences, as shown in Fig~\ref{fig:analysis}.

For trajectories, real videos show continuous and smooth variations over time, forming stable and coherent geometric paths. In contrast, generated videos display irregular behavior: the trajectories jitter, drift, or abruptly change direction. 
For EMA-residuals, taking the angular residual as an example, real videos produce low-magnitude and temporally stable residual curves. However, generated videos show spikes and high-frequency fluctuations across the residual channels. These observations demonstrate that Grab-3D effectively leverages 3D geometric temporal consistency to detect generated videos.

\begin{table}[t]
\centering
\caption{Ablation study on geometric modules testing on cross-domain. GPE: geometric positional encoding; GA: geometry attention branch; EMA: EMA-based geometric residuals. Metrics are overall averages on the evaluation split.}
\label{tab:ablation}
\small
\setlength{\tabcolsep}{4pt}
\renewcommand{\arraystretch}{1.0}
\begin{tabular}{l|cc|cc|cc}
\toprule
\cmidrule(lr){2-3} \cmidrule(lr){4-5} \cmidrule(lr){6-7}
&  AUC & $\Delta$ & F1 & $\Delta_{\text{F1}}$ & AP & $\Delta_{\text{AP}}$ \\
\midrule
$\textit{w/o}$ GPE      &  94.86 & -0.29 &  79.29 & -5.58 & 93.93 & -1.48 \\
$\textit{w/o}$ GA       &  93.83 & -1.32 &  82.13 & -2.74 & 94.01 & -1.40 \\
$\textit{w/o}$ EMA      &  93.98 & -1.17 &  79.79 & -5.08 & 94.69 & -0.72 \\
\textbf{\textit{Grab-3D (ours)}} & \textbf{95.15} & \textbf{-} &  \textbf{84.87} & \textbf{-} & \textbf{95.41}& \textbf{-} \\
\bottomrule
\end{tabular}
\end{table}

\subsection{Ablation Study}
In this section, we study how useful each proposed module is. Four experimental setups are evaluated: ``$\textit{w/o}$ GPE'', which degrades the geometric PE with sinusoidal PE; ``$\textit{w/o}$ GA'', which disables the geometric attention branch; ``$\textit{w/o}$ EMA'', which removes the EMA-based geometric residuals from the classifier.

As shown in Table~\ref{tab:ablation}, removing any component leads to a noticeable performance drop across all metrics.
Among them, GPE has the largest influence on F1 (-5.58\%), indicating that encoding positional information through geometric differences is effective for modeling temporal order.
Disabling the geometric attention branch (GA) causes a decline in both AUC and AP, showing that modeling inter-frame geometric correlations serves as an effective complement to temporal attention.
Finally, removing the EMA-based residuals results in a 5.08\% decrease in F1, demonstrating that capturing dynamic geometric instability through temporal residuals can improve classification accuracy.


\section{Conclusion}
\label{sec:conclusion}

This paper introduces 3D geometric temporal consistency into AI-generated video detection and proposes a geometry-aware temporal framework for detecting AI-generated videos (Grab-3D).
A large-scale video dataset of static scenes has been developed, comprising 3,322 real-generated video pairs that cover diverse scenes and generation models.
We propose a geometric-aware transformer equipped with geometric positional encoding, temporal-geometric attention, and geometric classifier head, achieving state-of-the-art performance, as demonstrated by extensive benchmarks. 
Ablation study highlights the effectiveness of each geometric-aware module, and qualitative analysis demonstrates the underlying 3D geometric differences. This work is expected to inspire further research on 3D geometric consistency in AI-generated video detection.

{
    \small
    \bibliographystyle{ieeenat_fullname}
    \bibliography{main}
}

\clearpage
\setcounter{page}{1}
\maketitlesupplementary
\appendix

\section{Additional Experiment Settings}
\label{sec:impl}

We describe additional training details, hyperparameters, and geometry-head pretraining settings that are omitted from the main paper.

\paragraph{Main Training Details}

All of our models are trained on 3 $\times$ NVIDIA RTX A6000 GPUs. For pre-processing, each video is randomly cropped to a 5-second segment, from which 80 frames are uniformly sampled and set in JPG format. During training, we use cross-entropy loss to optimize the classifier head and the AdamW optimizer with an initial learning rate of $10^{-4}$ and weight decay of $10^{-4}$ with a cosine annealing schedule and 5-epoch warm-up. We apply label smoothing of 0.02, dropout of 0.1, and gradient clipping with a max norm of 1.0 to stabilize optimization.

\paragraph{Geometry-head Pretraining Settings}
The geometric head is pre-trained exclusively on the real portion of our static video dataset, using only vanishing-point features and line segments. This stage aims to learn stable geometric constraints that are later used to compute EMA-based residuals during main training. We freeze the backbone and optimize only the geometric head for 30 epochs using AdamW with a learning rate of $2\times10^{-4}$ and weight decay of $10^{-4}$. 

\section{Data processing}
\begin{figure*}[t]
  \centering
  \includegraphics[width=1.0\linewidth]{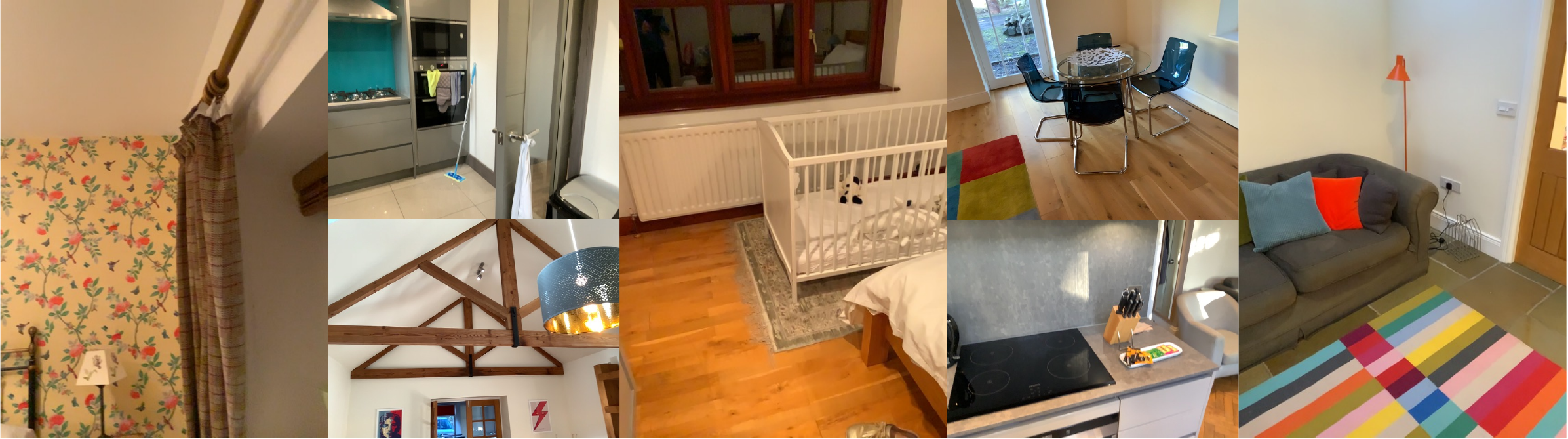}
  \caption{Example real-world samples from our collected video dataset of static scenes, covering diverse scenarios, viewpoints, and geometric layouts.}
  \label{fig:real}
\end{figure*}

\begin{figure*}[t]
    \centering
    \begin{subfigure}{\linewidth}
      \centering
      \includegraphics[width=1.0\linewidth]{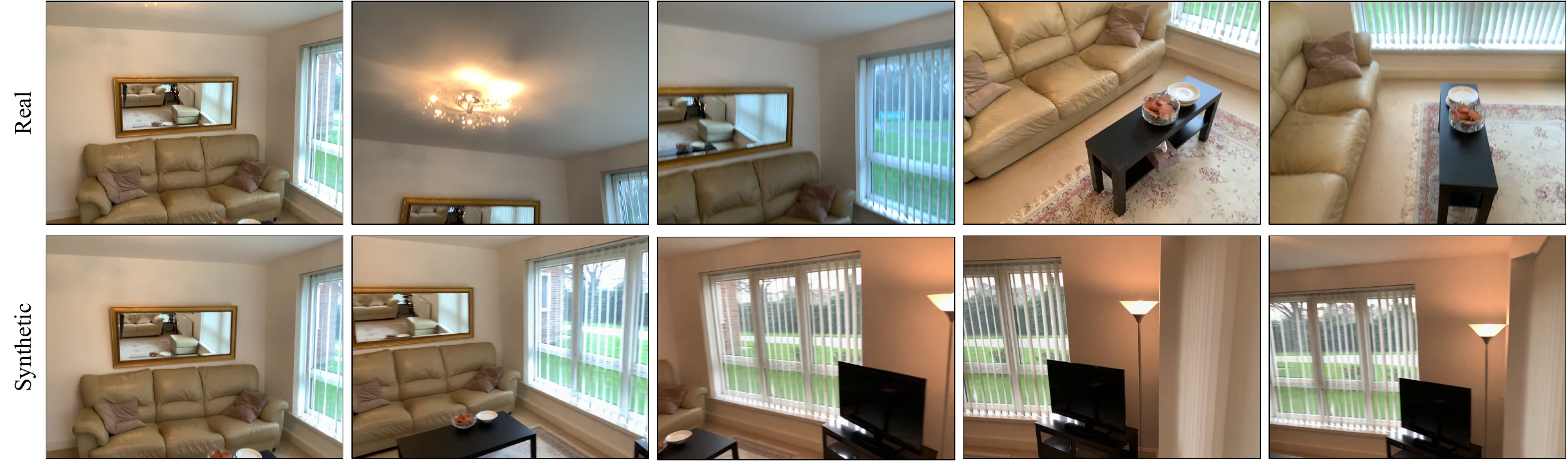}
      \caption{Synthetic video is generated by \textbf{Sora} with the prompt: \textit{Static indoor scene with no people, no object movement, only camera moving. The living room features a beige leather sofa with two cushions, positioned against a wall with a large mirror above it. A black coffee table sits in front of the sofa, holding a bowl of fruit and a plate. The room has a patterned rug under the table and a light-colored carpet covering the rest of the floor. A window with vertical blinds is to the right, allowing natural light to filter through. A flat-screen TV is mounted on a stand near a glass door. The scene should be completely still with no moving objects, no people walking or moving around. The camera moves smoothly to explore the whole space.}}
      \label{fig:6a_sora}
    \end{subfigure}
    \hfill
    \begin{subfigure}{\linewidth}
      \centering
      \includegraphics[width=1.0\linewidth]{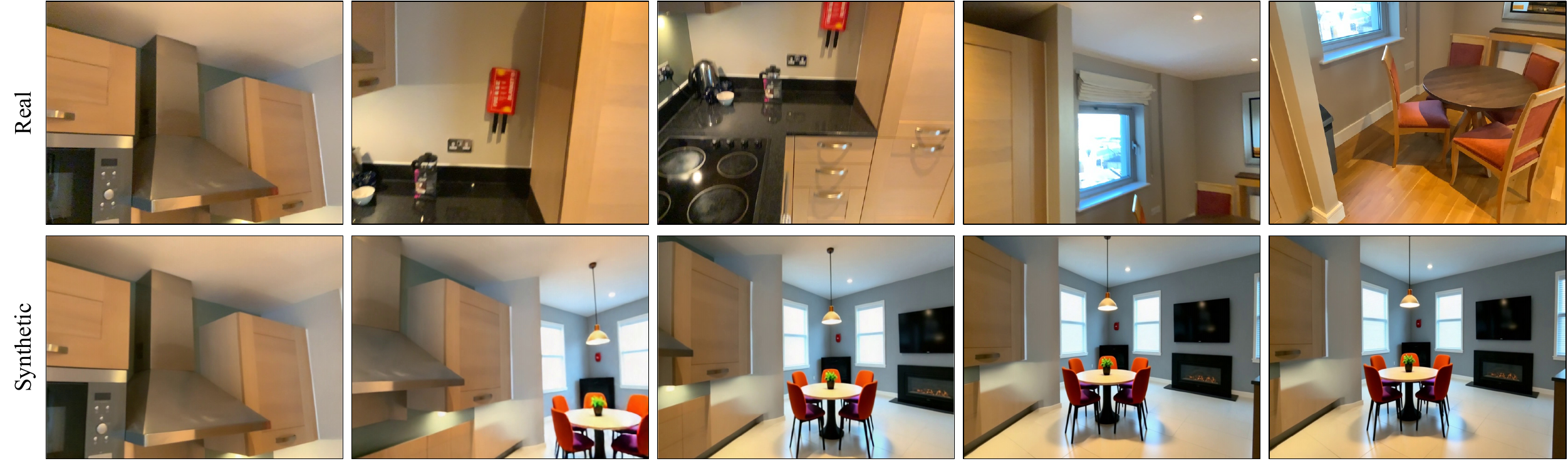}
      \caption{Synthetic video is generated by \textbf{Runway Gen-4} with the prompt: \textit{Static indoor scene with no people, no object movement, only camera moving. The indoor scene is a modern kitchen and dining area. The kitchen has light wooden cabinets, a black countertop with a stove, and a red fire alarm mounted on the wall. The floor is tiled in white. The dining area features a round wooden table with four chairs, each with a mix of orange and purple cushions. The room has a light-colored ceiling with recessed lighting. The overall color scheme includes neutral tones with pops of color from the furniture. The scene should be completely still with no moving objects, no people walking or moving around. The camera moves smoothly to explore the whole space.}}
      \label{fig:6b_gen4}
    \end{subfigure}
    \caption{Examples of paired real–synthetic video generation from different models.}
    \label{fig:paired1}
\end{figure*}

\begin{figure*}[t]
    \centering
    \begin{subfigure}{1.0\linewidth}
      \centering
      \includegraphics[width=1.0\linewidth]{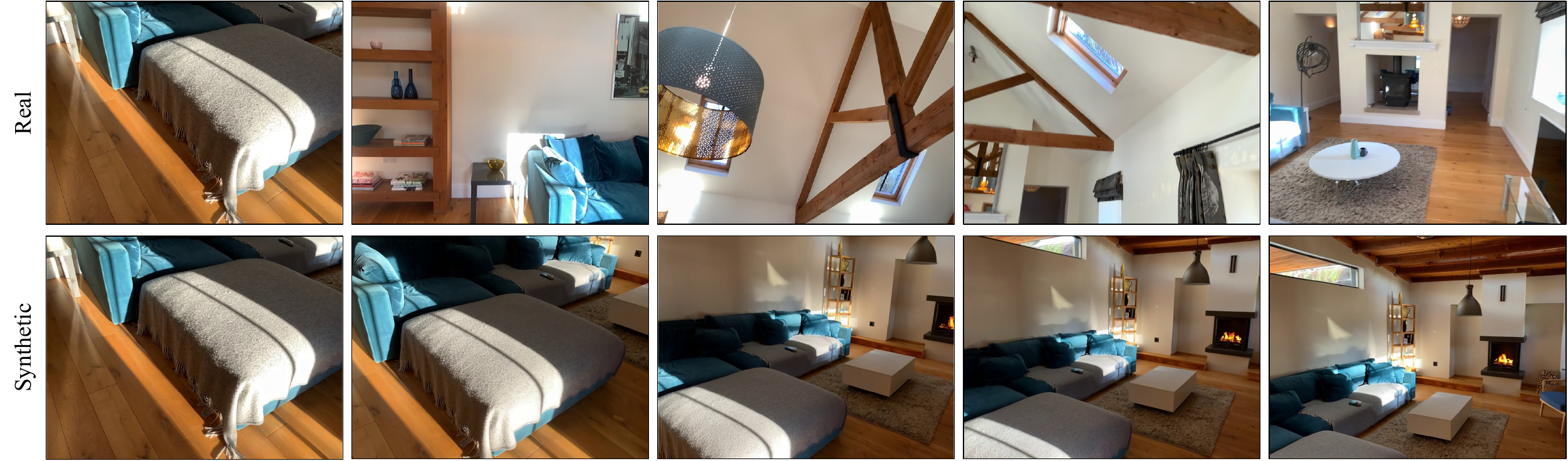}
      \caption{Synthetic video is generated by \textbf{Wan2.2} with the prompt: \textit{Static indoor scene with no people, no object movement, only camera moving. The living room features a light blue sofa with dark blue cushions and a gray throw blanket. A small side table holds decorative items near a bookshelf filled with books. The ceiling has exposed wooden beams, a hanging lamp, and a skylight. The floor is wooden, and a white coffee table sits on a rug in front of a fireplace with a modern design. The space is well-lit with natural light streaming through the windows and skylights. The scene should be completely still with no moving objects, no people walking or moving around. The camera moves smoothly to explore the whole space.}}
      \label{fig:wan}
    \end{subfigure}
    \hfill
    \begin{subfigure}{1.0\linewidth}
      \centering
      \includegraphics[width=1.0\linewidth]{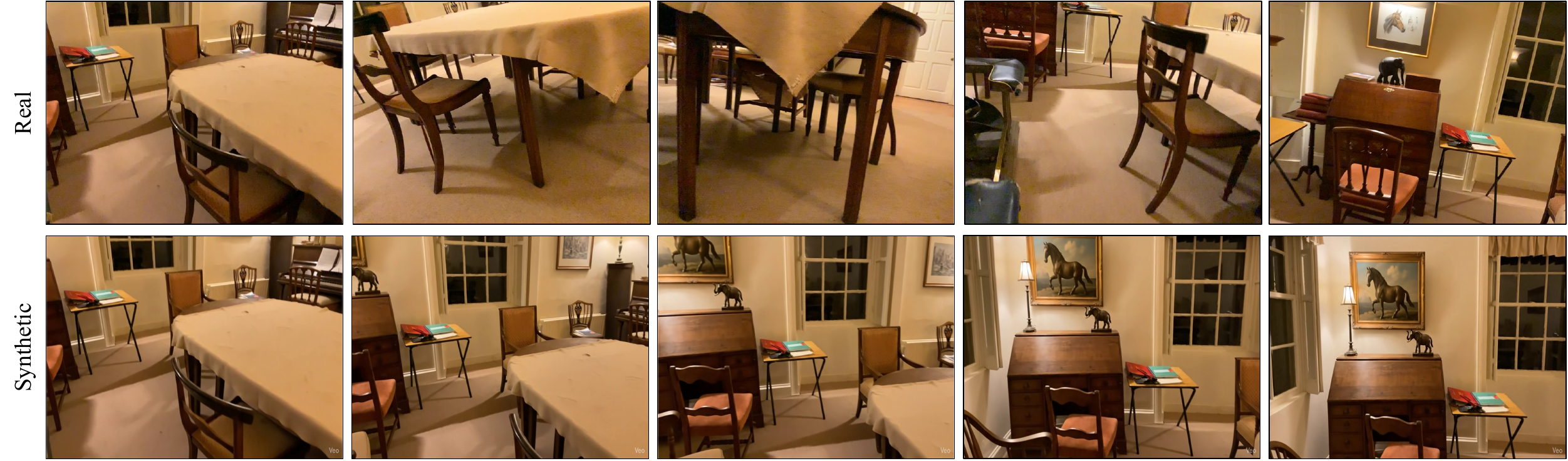}
      \caption{Synthetic video is generated by \textbf{VEO} with the prompt: \textit{Static indoor scene with no people, no object movement, only camera moving. The room is dimly lit with warm tones. A large wooden dining table covered in a white cloth occupies the center, surrounded by matching chairs. A small folding table holds books near a window with grid panes. A wooden desk with an elephant figurine sits against the wall, accompanied by a chair. The floor is carpeted, and a framed horse portrait hangs above the desk. The overall layout suggests a cozy, lived-in space. The scene should be completely still with no moving objects, no people walking or moving around. The camera moves smoothly to explore the whole space.}}
      \label{fig:veo}
    \end{subfigure}
    \hfill
    \begin{subfigure}{1.0\linewidth}
      \centering
      \includegraphics[width=1.0\linewidth]{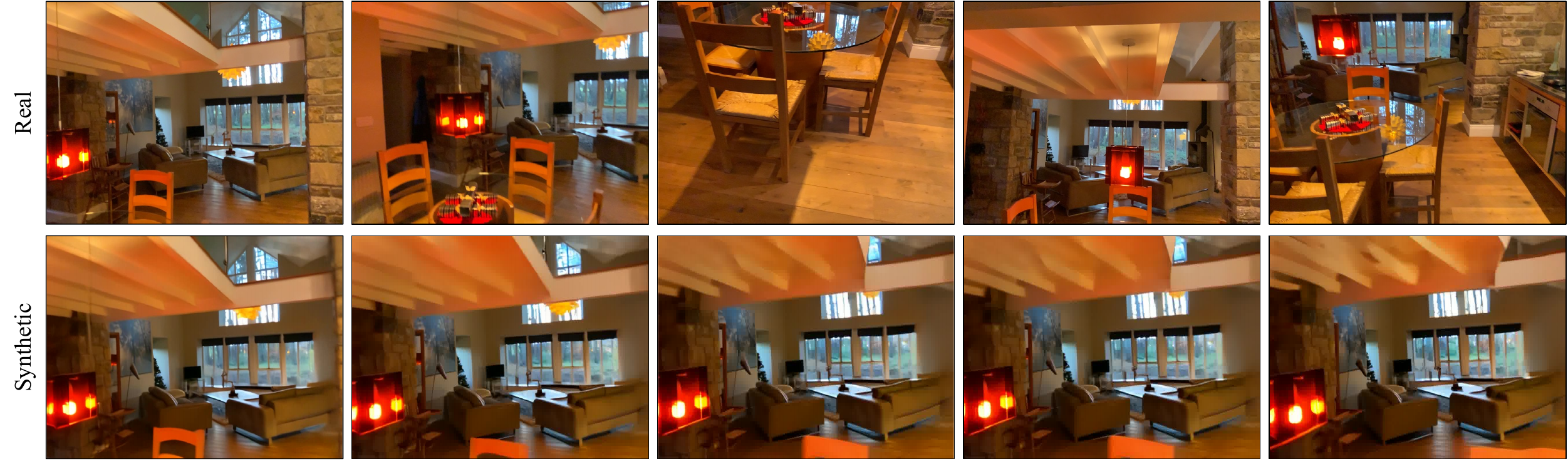}
      \caption{Synthetic video is generated by \textbf{SVD} with an image-only prompt (SVD does not support text input, only image conditioning).}
      \label{fig:svd}
    \end{subfigure}
    \caption{Examples of paired real–synthetic video generation from different models.}
    \label{fig:paired2}
\end{figure*}

We collect real videos as described in Sec.~\ref{sec:staticvideodataset}, including samples from Arkitscenes~\cite{baruch2021arkitscenes} and NYUdepth~\cite{Silberman:ECCV12}, totaling 2,284 videos. To ensure consistent content descriptions for generating paired synthetic samples, we slice each real video into equal-length 5-second clips, as shown in Fig.~\ref{fig:real}. Additionally, we extract the first frames of these video clips as the image prompt. For the video description, we feed each real clip into the large language model (\textit{Qwen}~\cite{bai2025qwen2}) together with a restrictive prompt~\textit{“Provide a description of this scene. Focus on describing the layout, objects, and architectural features. Describe only the static scene and spatial arrangement, ignore any movement.”}. We then combine the descriptions with the first frames and feed them into each video generation model to produce the synthetic videos. During this process, we further append a static-scene constraint to the textual prompt:\textit{"Static indoor scene with no people, no object movement, only camera moving. \{description\} The scene should be completely still with no moving objects, no people walking or moving around. The camera moves smoothly to explore the whole space."}. More AI-generated samples are shown in Fig.~\ref{fig:paired1} and Fig.~\ref{fig:paired2}.

\section{Detection with Foreground Masks}
\begin{figure*}[t]
  \centering
  \includegraphics[width=1.0\linewidth]{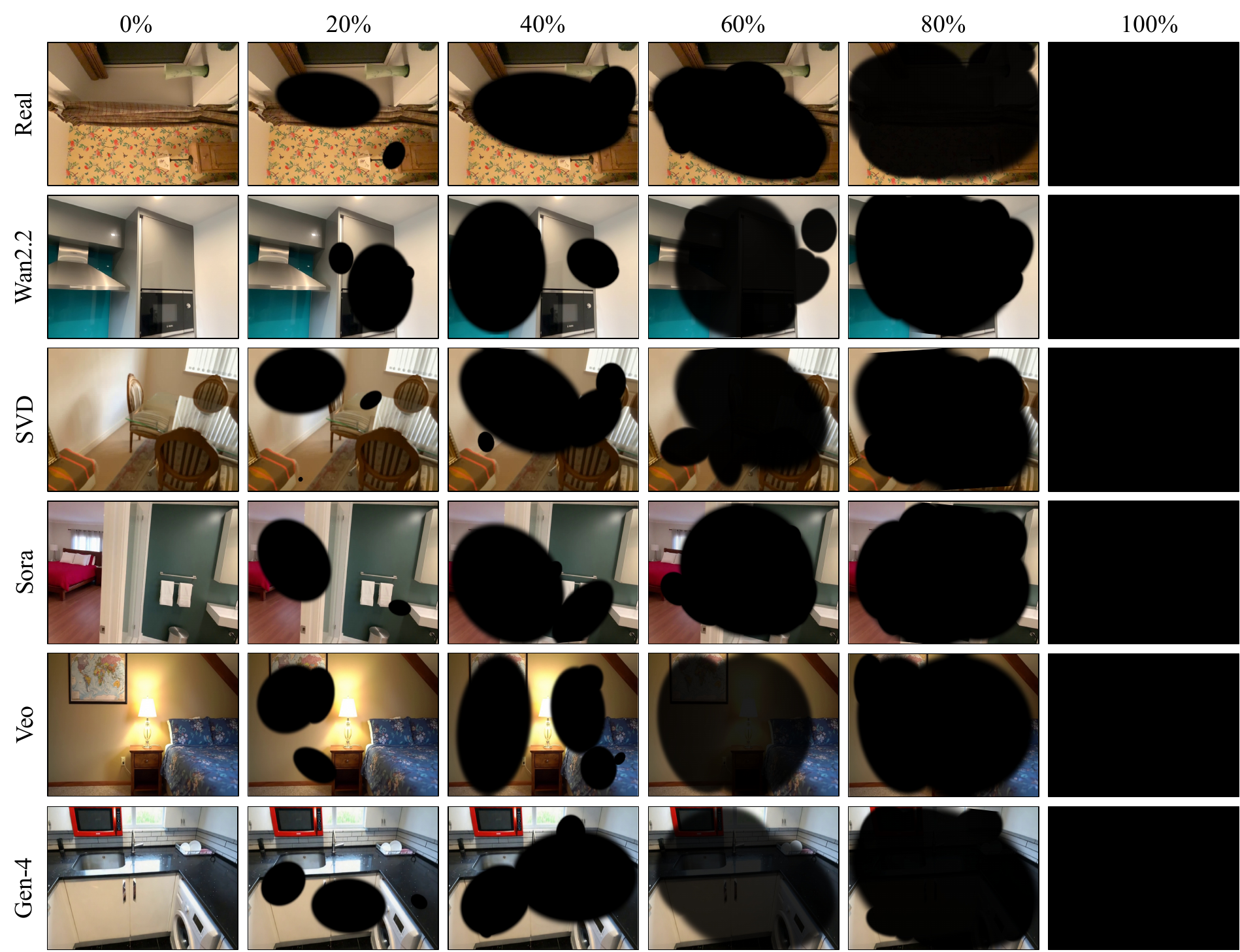}
  \caption{Examples of synthetic dynamic masks applied to real and AI-generated video frames at different occlusion ratios (0–100\%).}
  \label{fig:mask_fig}
\end{figure*}

We evaluate our model's generalizability on more general AI-generated videos. Inspired by SAM-2~\cite{ravi2024sam}, dynamic foreground can be separated from videos to obtain a static background. However, this process leaves masks on the video frames, which weakens the extraction of 3D features. To simulate this scenario, we block-mask video frames in the static video dataset with different occlusions (Fig.~\ref{fig:mask_fig}).

\begin{figure}[t]
  \centering
  \includegraphics[width=1.0\linewidth]{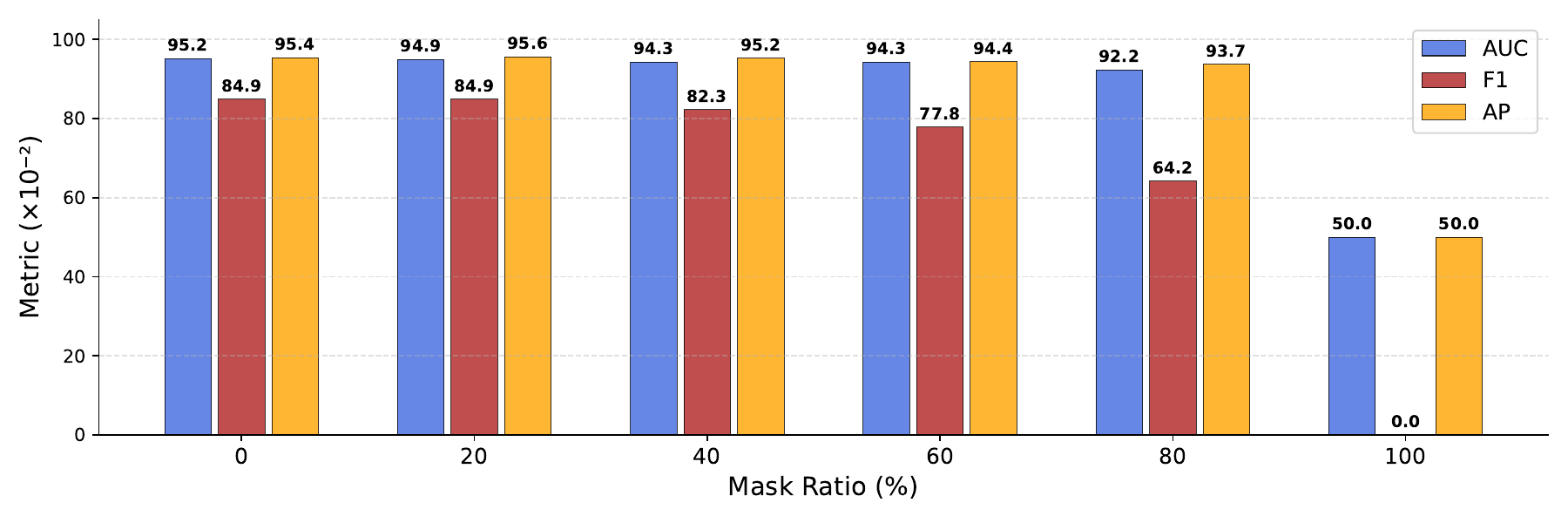}
  \caption{Evaluation of Grab-3D under synthetic dynamic masks. Increasing portions of each frame (0–100\%) are block-masked to simulate occlusions left by foreground removal.}
  \label{fig:mask}
\end{figure}

As shown in Fig.~\ref{fig:mask},  Grab-3D remains highly robust when up to 40–60\% of each frame is occluded. The AUC and AP curves exhibit only slight degradation, indicating that the model primarily relies on globally consistent geometric cues that remain visible despite partial blockage. 
Performance drops notably once the masking ratio exceeds 60\%. In these conditions, large parts of the scene geometry are removed, especially line segments that contribute to stable vanishing-point estimation, resulting in reduced reliability of the geometric features. When masking reaches 100\%, no geometric information remains, and the model degenerates to random prediction, as expected. It is worth noting that extreme occlusion levels of 60–80\% rarely occur in general videos, suggesting that extending Grab-3D to general AI-generated video detection could be a potential future work.

\section{Limitations}
Although this work advances 3D geometric consistency in AI-generated videos, several challenges remain:
1) Our dataset has limitations in scale, video length, and diversity, which restrict the exploration of geometric patterns in real-world videos.
2) The generalization of Grab-3D to dynamic and general AI-generated videos requires further validation.

\end{document}